\documentclass{article}
\usepackage{spconf,amsmath,graphicx,multirow}
\usepackage{times}
\usepackage{textcomp}
\usepackage{latexsym}
\usepackage{amssymb,mathtools}
\usepackage{graphicx,paralist,multirow}
\usepackage{url}

\DeclareMathOperator{\R}{\mathbb{R}}

\title{Go Beyond Plain fine-tuning: Improving Pretrained Models For Social Commonsense}
\name{Ting-Yun Chang$^1$\sthanks{Work was done while the first and fifth author were interns at Amazon}, Yang Liu$^2$, Karthik Gopalakrishnan$^2$, Behnam Hedayatnia$^2$,
 {Pei Zhou$^3$, Dilek Hakkani-T\"ur$^2$}}
\address{$^1$ Academia Sinica, Taiwan; $^2$ Amazon, Alexa AI, USA; $^3$ 
University of Southern California}
\begin{document}
\maketitle
\begin{abstract}
Pretrained language models have demonstrated outstanding performance in many NLP tasks recently. However, their social intelligence, which requires commonsense reasoning about the current situation and mental states of others, is still developing. Towards improving language models' social intelligence, in this study we focus on the Social IQA dataset, a task requiring social and emotional commonsense reasoning.
Building on top of the pretrained RoBERTa and GPT2 models, we propose several architecture variations and extensions, as well as leveraging external commonsense corpora, to optimize the model for Social IQA.
Our proposed system achieves competitive results as those top-ranking models on the leaderboard.
This work demonstrates the strengths of pretrained language models, and provides viable ways to improve their performance for a particular task. 
\end{abstract}
\section{Introduction}
Recently, there has been emerging research interest in developing machine common sense. Several commonsense datasets have been proposed~\cite{talmor2019commonsenseqa,huang2019cosmos,zellers-etal-2019-hellaswa,sap2019socialiqa,sakaguchi2019winogrande}.
In particular, Social IQA was introduced in \cite{sap2019socialiqa}, which is a multiple-choice QA dataset probing machine's emotional and social commonsense reasoning in a variety of everyday situations.
Afterward, different efforts have been put forth for this challenging task.  
One line of work introduces additional knowledge bases, providing the model with more information underlying the given text~\cite{shwartz2020unsupervised,lin2019kagnet,mitraadditional, paul2020social}. 
Another line generalizes existing well-performed models, focusing on zero-shot setting~\cite{shwartz2020unsupervised, bosselut2019dynamic}.
Generally, large pretrained language models are leveraged in previous work, such as BERT~\cite{devlin2019bert}, RoBERTa~\cite{liu2019roberta}, GPT and GPT2~\cite{radford2018improving, radford2019language}, and  T5~\cite{raffel2019exploring}.
For example, \cite{khashabi2020unifiedqa} recently reformulates Social IQA as a generation task and achieves remarkable performance by fine-tuning T5~\cite{raffel2019exploring}. 

A standard practice in NLP recently is to simply fine-tune the large pre-trained language models on the particular downstream task. 
This is indeed a very strong baseline for many tasks, some even reaching human-level performance.
It is also the backbone of most of the top-performing methods for the Social IQA task. 
In this study, we start from such a practice, and aim to improve it in the context of Social IQA from several directions, which we believe are applicable to other language understanding tasks as well. 
\begin{enumerate} 
\item
(i) How can we improve the model's stability? Training instability is a common issue when fine-tuning large pretrained language models on downstream tasks~\cite{devlin2019bert, dodge2020fine}.
We propose to use multi-task learning with both generation loss and classification loss. We also introduce an ensemble method for incorporating different pretrained language models, such as GPT2 and RoBERTa, making the final result less sensitive to an individual model. 
\item
(ii) How can we optimize the model for the multi-choice task?  Rather than treating each answer candidate independently, we propose to look at all the choices together, and thus allowing the model to compare them when making predictions.  In addition, we introduce inter-segment attention to model the relationships between the given context, question, and answer options for the Social IQA multi-choice task.
\item
(iii) How can we utilize other datasets? There are other multiple-choice datasets related to common sense, with different focuses from Social IQA.  We adopt a two-phase approach, where we first fine-tune the models on other tasks, and then on Social IQA. This way the model learns general common sense first and then is optimized for the specific Social IQA task. 
\end{enumerate}

We conduct thorough experiments on the Social IQA task, showing that while the model size and pretraining corpora have a great influence on the final result, sophisticated architecture and similar-domain knowledge usage in the fine-tuning step are very beneficial.
Our experimental results demonstrate the contribution and the competitive performance of different methods we proposed. 
We expect that our analyses may help future work to better understand pretrained models' capability of social intelligence.

\section{Using Pretrained Language Models as Baseline}
\label{sec:problem_form}
\begin{figure}[ht]
\centering
\includegraphics[width=8cm]{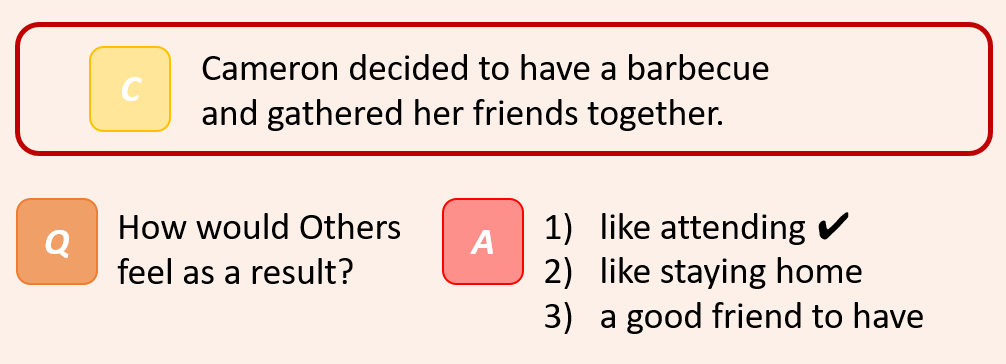}
\caption{An example in the Social IQA data.}
\label{fig:example}
\end{figure}

In Social IQA~\cite{sap2019socialiqa}, given a context $C$ about an event, and a corresponding question $Q$, the goal is to select the correct answer from the set $A=(A_1, A_2, A_3)$. Figure~\ref{fig:example} shows an example.

A typical approach for solving this kind of multiple-choice problem with a pretrained transformer-based language model is done by concatenating $C$, $Q$, and $A_i$ with a separator token, and then letting the model output a score via an MLP (multi-layer perceptron) built on top of the final hidden representation of the classifier token $[CLS]$. Finally, scores for each data point are normalized across all $(C, Q, A_i)$ instances with softmax.  
For example, if we are fine-tuning BERT, the input is formatted as below:
\begin{equation}
[CLS] \  C \  [SEP] \  Q \  [SEP] \  A_i \  [SEP] \nonumber
\end{equation} On the other hand, when using GPT-based models, the $[CLS]$ token is placed at the end of the sequence instead of the beginning, since it is feed-forward.
Cross-entropy loss is generally used for model training. 

As RoBERTa~\cite{liu2019roberta} has shown competitive performance on the Social IQA leaderboard, we use it as the main baseline model.  
In the following, we propose methods to improve the model's stability and better represent the multi-choice task. 

\section{Improving Model Robustness}
\subsection{Training Stability}
\label{sec:lm}
Training instability is a common issue when fine-tuning large pretrained language models on downstream tasks~\cite{devlin2019bert, dodge2020fine}. With a poor hyper-parameter setting, a model achieving state-of-the-art may perform no better than random guessing. A probable explanation is that the model suffers from \emph{catastrophic forgetting}~\cite{kirkpatrick2017overcoming, chen2019catastrophic}. 

To address this problem, in addition to cross-entropy loss for multiple-choice (MC) classification, we add the masked language modeling (MLM) loss~\cite{devlin2019bert} in a multi-task training setup. 
Therefore the loss in BERT-based models is the sum of the MLM and MC CE loss, and for GPT-based models, it is the unidirectional LM loss plus MC loss. 
The idea is that while a new downstream task might be challenging to learn at the beginning of the fine-tuning process, the generation loss can provide a stable objective that matches the pretraining phase. Therefore, the model could make steady progress and be less sensitive to different hyper-parameters.
Note that we only train the (masked) LM loss on inputs with the correct answer $A_{y^*}$, where $y^*$ is the reference label.

\subsection{Combining RoBERTa with GPT2}
In our initial experiments, we observed RoBERTa-large outperforms GPT2-medium model (both fine-tuned on the social IQA data set), with $\sim10\%$ performance gap.
With such a large difference, simply combining the two models' predictions is not an effective way for system combination. 
Since GPT2 is a unidirectional regressive model and there are some differences in the pretraining corpora used between GPT2 and RoBERTa, we expect these two models may be complementary. 
To fuse these two models effectively, we do not train them jointly since it is computationally expensive; instead, we propose a two-step approach, which only needs to keep one pretrained model in the memory at each step.

\begin{figure}[h]
  \centering
  \includegraphics[width=1.0\linewidth]{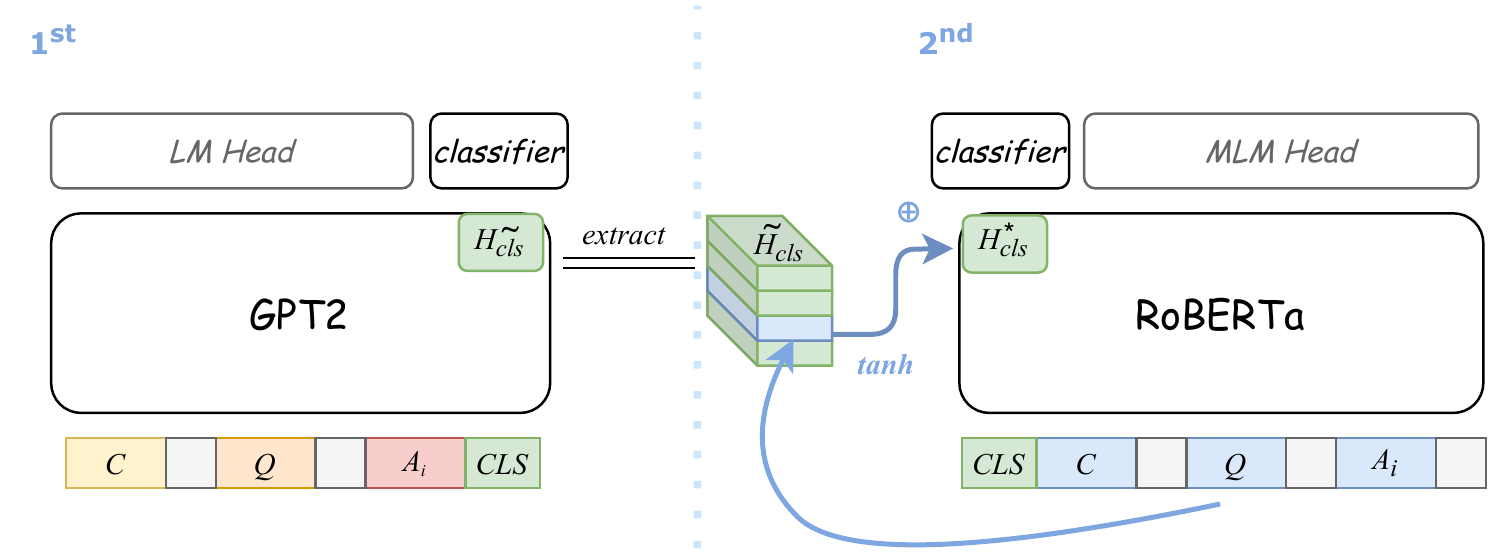}
  \caption{The pipeline of combining GP2 and RoBERTa. First, we fine-tune GPT2 and then store its representations for all the instances. Secondly, RoBERTa retrieves the corresponding one ($\Tilde{H_{CLS}}$) to use in its fine-tuning step.} 
  \label{fig:gpt2}
\end{figure}

Figure~\ref{fig:gpt2} illustrates the two steps. 
First, we fine-tune the GPT2 model on Social IQA, with the $[CLS]$ token placed at the end of the concatenated sequence of $C, Q$ and $A_i$, and an MLP classifier stacked upon it. 
We believe that the final hidden representation of $[CLS]$ condenses some semantic meanings of the entire sequence, and it provides useful information for this multi-choice task after fine-tuning. Hence, we extract this $[CLS]$ token representation and use it as an additional feature when we move on to fine-tune RoBERTa (the second step). Our proposed 2-step ensemble method is especially tailored to large pretrained models, while sharing a similar idea to the traditional Stacked generalization~\cite{wolpert1992stacked}.

Specifically, let $\Tilde{H_{CLS}}$ and $H_{CLS}^*\in {\R}^{d}$ be the $[CLS]$ hidden representation of GPT2 and RoBERTa respectively. For each training instance, we extract $\Tilde{H_{CLS}}$ from the fine-tuned GPT2, and then combine it with $H_{CLS}^*$ while fine-tuning RoBERTa for the multi-choice task:
\begin{align}
\Bar{H_{CLS}} &= [tanh(\Tilde{H_{CLS}}) \oplus H_{CLS}^*] \\
Score &= \Bar{H_{CLS}}\ W_t + b_t
\end{align}
where $\oplus$ can be concatenation or mean-pooling \footnote{Note that in Hugging Face's implementation, BERT-based models' final $[CLS]$ representation is normalized by tanh, so we scale $\Tilde{H_{CLS}}$ by tanh as well before the combination. RoBERTa-large and GPT2-medium have the same hidden dimsention $d=1024$.}. 
Finally, the combined $\Bar{H_{CLS}}$ is passed to RoBERTa's classifier to predict the score. In both steps, we apply the generation loss besides classification loss (Section~\ref{sec:lm}), using unidirectional LM for GPT2 and MLM for RoBERTa.
Note that $\Tilde{H_{CLS}}$ serves as external features for RoBERTa and will not be tuned while updating RoBERTa's parameters.

\vspace{-0.1in}
\section{Other Proposed Model Variations}

\subsection{Cross-Segment Attention}
\label{sec:multiway}
In the standard self-attention mechanism, each token attends to \emph{all the tokens} in the sequence. Yet in a task like Social IQA, there is a distinct separation of context, question, and answer, and we would like to emphasize the differences between these segments. 
Previous work has dealt with this property with different architectures, usually using complex cross-segment (or named multiway) attention blocks ~\cite{huang2019cosmos,Seo2017Bidirectional, zhu2018hierarchical, Wang2018YuanfudaoAS}. 

In this paper, we apply a simple adaption of previous work~\cite{huang2019cosmos} that suits BERT well. We keep the $[SEP]$ token in the input and leave the self-attention layer untouched, but stack an inter-segment attention layer using the final hidden representations.
This additional attention layer attends between different segments only. 
This design is based on two considerations: 
\vspace{-0.1in}
\begin{itemize}
    \item We would like to follow the original BERT framework that inherits the huge transformer block pretrained on large corpora, and fine-tune it with the simple task-specific layer built atop~\cite{devlin2019bert}. 
\vspace{-0.1in}
    \item We want to keep the advantage of seeing global information through self-attention~\cite{vaswani2017attention}, at the same time stress the difference of segments on the semantic level, which is believed to be learned by the higher layers of BERT~\cite{tenney2019bert}.
\end{itemize}

\begin{figure}[ht]
\centering
\centering
\includegraphics[width=7.5cm]{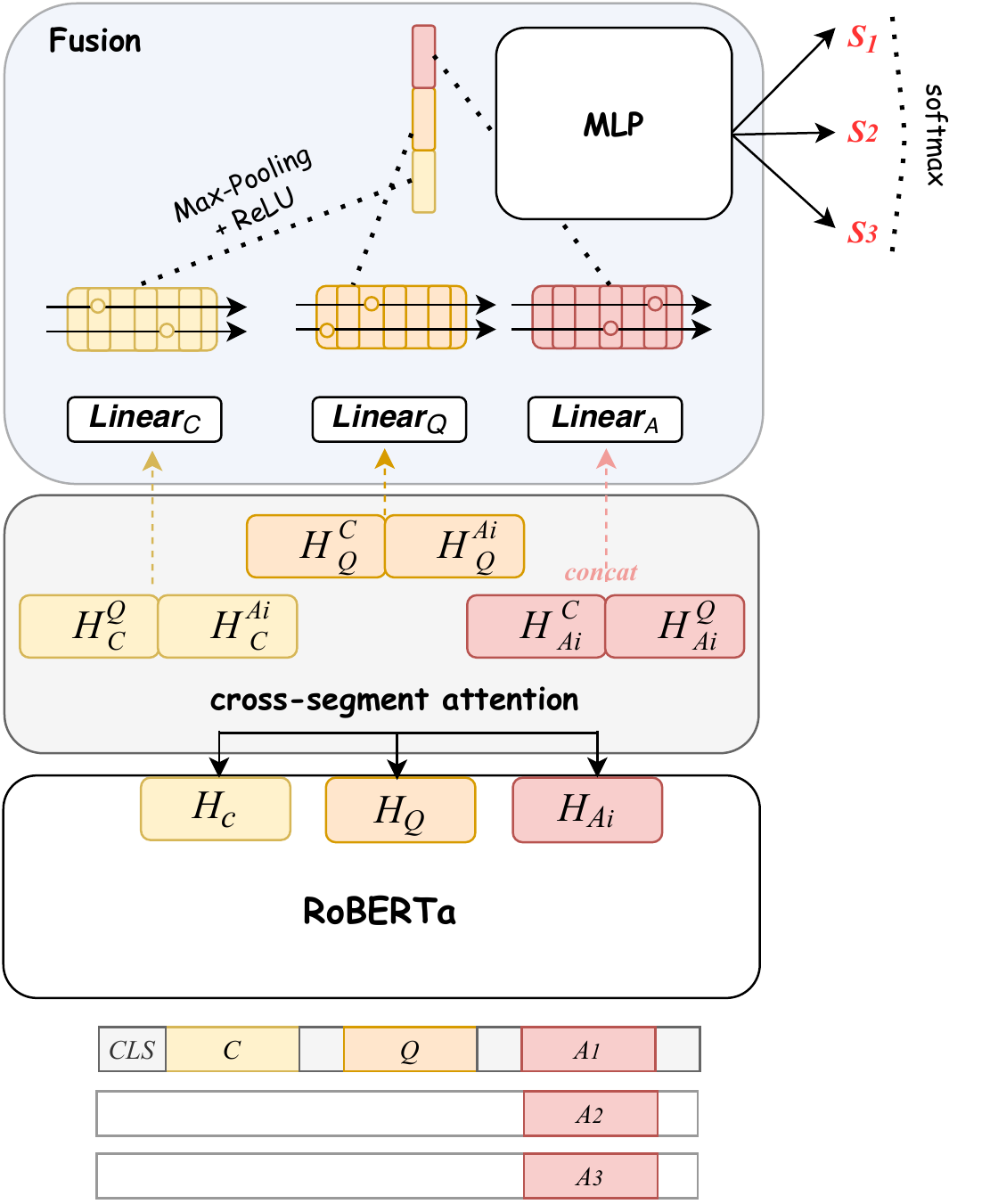}
\caption{The Cross-Segment Attention model.}
\label{fig:multiway}
\end{figure}

\vspace{-0.1in}
Formally, let $d$ be the hidden dimension, $l$ be the sequence length of the input, $H_C, H_Q$, and $H_{A_i}\in {\R}^{l\times d}$ be BERT's final representation (zero-padded to $l$ for parallelism), corresponding to the context, question and the $i^{th}$ answer respectively. We apply the inter-segment attention:
\begin{align} 
\hat{H_C} &= H_C\ W + \mathbf{1}*b_1^T,\nonumber \\
\hat{H_Q} &= H_Q\ W + \mathbf{1}*b_1^T, \nonumber \\
\hat{H_{A_i}} &= H_{A_i}\ W + \mathbf{1}*b_1^T, \nonumber \\
H_C^Q &= Softmax(\frac{\hat{H_C} \hat{H_Q}^T}{\sqrt{d}}) \ H_Q \nonumber \\
H_C^{A_i} &= Softmax(\frac{\hat{H_C} \hat{H_{A_i}}^T}{\sqrt{d}}) \  H_{A_i} \label{eq:attn}
\end{align}
where $W \in {\R}^{d\times d}$, $b \in {\R}^{d}$, $\mathbf{1} \in {\R}^{l}$ (a vector of all-ones), $H_C^Q$ and $H_C^{A_i}\in {\R}^{l\times d}$ are question-attentive and answer-attentive representation (zero-padded) for the context $C$ respectively, and $\sqrt{d}$ is used to smooth the softmax results~\cite{vaswani2017attention}. 
Similarly, we obtain context-attentive and answer-attentive representation ($H_Q^C$ and $H_Q^{A_i}$) for question $Q$, and context-attentive and question-attentive representation for answer $A_i$ ($H_{A_i}^Q$ and $H_{A_i}^C$).

Once we obtain all these representations ($H_C^Q$, $H_C^{A_i}...$), we can then apply a \emph{fusion layer} to utilize all of them and output a score for classification. 
In our experiments, we evaluated different commonly used fusion methods, including element-wise addition, subtraction, concatenation, ReLU and max-pooling over the sequence dimension, and several linear transformations.  The final model architecture is decided by our experimental results.
Figure~\ref{fig:multiway} illustrates the model using cross segment attentions. 

\subsection{Seeing all Choices}
In the baseline model introduced in Section~\ref{sec:problem_form}, the model assigns a score for each answer independently, without looking at other answer candidates.
However, we human often compare all choices when solving multiple-choice problems, and 
when multiple answers look reasonable, we choose the most plausible one, and we use the method of exclusion frequently.
Doing these kinds of reasoning requires comparisons of different choices, instead of independently generating prediction scores for each candidate. 
Motivated by this, we re-format the input to allow the model to see all the choices as below:
\begin{align}
[CLS]\ C \ [SEP]\ Q \ [SEP]\ A_2 \ [SEP]\ A_3\ [UNK]\  A_1 \nonumber \\
[CLS]\ C \ [SEP]\ Q \ [SEP]\ A_1 \ [SEP]\ A_3\ [UNK]\  A_2 \nonumber \\
[CLS]\ C \ [SEP]\ Q \ [SEP]\ A_1 \ [SEP]\ A_2\ [UNK]\  A_3 \nonumber
\end{align}
where the answer after the unknown token $[UNK]$ is the target candidate, i.e., the model's prediction score is for that choice.

\vspace{-0.1in}
\section{Using External Resources}
The models introduced in the previous section heavily rely on fine-tuning pretrained models on the Social IQA dataset. Given the size of the training data and the complexity of social common sense, it is likely the models are undertrained. 
We believe it is beneficial for the model to access external information to learn more about language related to social scenarios. There are plenty of common sense resources with different focuses~\cite{talmor2019commonsenseqa,huang2019cosmos,zellers-etal-2019-hellaswa,sakaguchi2019winogrande, bisk2020piqa, liu2004conceptnet, sap2019atomic}, which may be very helpful for our model to learn more broad and general concepts of our world. In \cite{chang2020incorporating},  we investigated how to incorporate knowledge graph into the pretrained models.

In this study, our focus is on leveraging other multiple-choice datasets of common sense to help the model learn more general concepts:
\begin{itemize}
\item HellaSwag~\cite{zellers-etal-2019-hellaswa} tests a model's ability to choose the most plausible continuation of a story. There are no questions in this data, so each instance is formatted as $[CLS] \ C \ [SEP] \ A_i \ [SEP]$.
\item  CosmosQA~\cite{huang2019cosmos} evaluates machine's reading comprehension of people's everyday narratives. It is formatted the same as Social IQA (see Section~\ref{sec:problem_form}).
\item WinoGrande~\cite{sakaguchi2019winogrande} improves the scale and the hardness of WSC~\cite{levesque2012winograd}, focusing on coreference resolution. WinoGrande contains one sentence with a blank and two options to be filled in. Following~\cite{sakaguchi2019winogrande}, an example is formatted as \emph{" [CLS] The trophy doesn't fit into the brown suitcase because the [SEP] \_ is too large. [SEP]"}, where the blank is filled with either $A_1$ or $A_2$.
\end{itemize}
Note that these additional datasets are built based on different sources, and ATOMIC~\cite{sap2019atomic}, the source of Social IQA, is not included, therefore using these datasets should not compromise testing on Social IQA's data.

We adopt 2-stage fine-tuning~\cite{phang2018sentence, pruksachatkun2020intermediate}, where we use these additional datasets in the first stage fine-tuning, and then continue to tune the model on Social IQA in the second stage. Specifically, in the first stage, we use a multi-task learning setup as MTDNN~\cite{liu2019multi}, where all the additional datasets are jointly trained at equal sample rates.
Since all the datasets are multiple-choice tasks, we adopt a unified classifier to assign a score for each choice, instead of one classifier per task. Empirically, we found no significant differences in sharing it or not. We trained our model with $N$-way cross-entropy loss, where $N$ is the number of choices, with $N_{SocialIQA}=3,\ N_{CosmosQA}=4,\ N_{HellaSwag}=4,$ and $ N_{WinoGrande}=2$. 
Note that we not only utilize the classification labels in these external datasets, but also train our model  with MLM loss (Section~\ref{sec:lm}) on the diverse text they provided.
Fine-tuning on these datasets is expected to improve the model's general representation for common sense related words and tasks. 
The idea is similar to ~\cite{gururangan2020don}, where they show the benefits when continuing to pretrain models on data in the task domain.

\section{Experiments and Results}
\subsection{Experimental Setup}
We train our models on the 33k Social IQA training instances, running hyper-parameters search for every proposed model respectively, and report their best performance on the dev set. We ran a grid search over the learning rate in $\{5e-6,\ 1e-5,\ 2.5e-5,\ 5e-5\}$, and the ~\emph{effective batch size} (number of GPUs $\times$ batch size per GPU $\times$ gradient accumulation steps) in $\{16,\ 32,\ 64,\ 256\}$. Every model was trained on \texttt{8 NVIDIA V100 16GB} GPUs, using Hugging Face's \texttt{transformers} toolkit~\footnote{\url{https://huggingface.co/transformers/}}.

\subsection{Results}
\textbf{Pretraining} To evaluate the impact of pretraining, we first fine-tuned several large pre-trained language models using the same framework as in \cite{sap2019socialiqa}, including variants of BERT, RoBERTa, and GPT2.  
Table~\ref{table:pretrain} shows the results. We can see a consistent trend: the larger the model size and the pretrained corpus, the better the performance is. 
Specifically, RoBERTa-large and GPT2-medium have similar model size; however, RoBERTa was pretrained on a larger corpus than GPT2's (160GB vs. 40GB). 
Furthermore, our experiment of training a GRU-RNN model from scratch only achieved about $52\%$ accuracy, again showing the necessity of using large-scale pretrained models for such a commonsense task.  
All our following experiments are based on RoBERTa-large as it performs the best among all these baseline models.

\begin{table}[ht]
  \centering
  \begin{tabular}{|l|c|c|}
  \hline
	\bf Model & \bf Accuracy (\%) & \bf Model Size (M) \\
  \hline\hline
    BERT-base & 63.3 & 110 \\
    BERT-large & 66.0 & 340 \\
    GPT2-medium & 66.2 & 345 \\
    GPT2-large & 71.7 & \bf{1558} \\
    RoBERTa-base & 69.0 & 125 \\
    RoBERTa-large & \bf{78.0} & 355 \\
  \hline
  \end{tabular}
  \caption{Results on Dev set using different pretrained models fine-tuned on Social IQA.}
  \label{table:pretrain}
\end{table}

\textbf{MLM} We then demonstrate the effectiveness of jointly training the MLM loss (see Section~\ref{sec:lm}) in Table~\ref{table:MLM}. As our goal is to improve training stability, we perform two rounds of 18 experiments with different learning rates and gradient accumulation steps, comparing the average performance of RoBERTa-large with and without MLM. While the best and worst performances of the two models are about the same, the average performance of the one with MLM is much better ($\sim5\%$ improvement). Note here we do not apply other proposed methods and follow the basic framework~\cite{sap2019socialiqa} to evaluate the impact of adding MLM in training.

\begin{table}[ht]
  \centering
  \begin{tabular}{|l|c|c|c|}
  \hline
	\bf Model & \bf Best & \bf Worst & \bf Mean, std\\
	\hline\hline
	RoBERTa-large & 78.0 & 32.9 & 63.8, 0.17 \\
	+ MLM & 78.4 & 33.2 & \underline{69.4}, 0.13\\
  \hline
  \end{tabular}
  \caption{Results (accuracy in \%) on the Dev set showing the effectiveness of incorporating MLM loss.}
  \label{table:MLM}
\end{table}

\textbf{Proposed Methods} We evaluate our proposed models and report their best performances on the dev set in Table~\ref{table:proposed}. All of the proposed methods lead to better results. Specifically, we experimented with three GPT2-medium models fine-tuned with different hyper-parameters as feature extractors, and found all of them are beneficial. However, when adopting features from GPT2-xl, which is $\sim5\%$ better than GPT2-medium on the Social IQA dev set, we found no improvement than using RoBERTa only. A possible explanation is that GPT2-medium's features are more complementary to RoBERTa-large. 
It is worth noting that when we applied such a GPT2-medium and RoBERTa-large combination to other commonsense datasets, we also found noticeable improvements in preliminary experiments, including CosmosQA, HellaSwag, and WinoGrande.
\begin{table}[ht]
  \centering
  \begin{tabular}{|l|c|}
  \hline
	\bf Model & \bf Accuracy (\%) \\
  \hline\hline
    Sap et al.~\cite{sap2019socialiqa} & 78.0 \\
    \hline
    See all Choices & 78.9 \\
    Cross-Segment Attention & 79.2  \\
    RoBERTa + GPT2 & 78.6, 79.1, 79.7 \\
    Multi-Choice Datasets & 79.3 \\ \hline
    Ensemble & 81.1 \\
  \hline
  \end{tabular}
  \caption{Comparisons of different proposed models. The three values in RoBERTa + GPT2 row correspond to three different GPT2-medium models fine-tuned on Social IQA.}
  \label{table:proposed}
\end{table}

 The last row in Table~\ref{table:proposed} shows the results when ensembling 5 models with majority vote, including See all Choices, Cross-Segment Attention, RoBERTa + GPT2, Multiple-Choice Datasets, and one system using external concept knowledge graph that has a classification accuracy of 79.2\% \cite{chang2020incorporating}.  This ensemble outperforms all the individual models. 
We also submitted this method for the test set to Social IQA leaderboard. 
Our system achieves $78.3\%$, a competitive result to other state-of-the-art models. 


\subsection{Analyses}
To better understand the behavior of pretrained models on the Social IQA task, we analyzed the RoBERTa-large model (with MLM) from several aspects.

\textbf{Few-Shot Learning}
We randomly chose 10\% training instances and found the model achieved $72.3\%$ on the dev set, which is better than any other models in Table~\ref{table:pretrain}. 
This may imply that a large pretrained model like RoBERTa does learn some commonsense of social intelligence during pretraining, and it only needs a small amount of training instances to adapt to a specific task. 

\begin{figure*}[ht]
  \centering
  \includegraphics[width=0.9\linewidth]{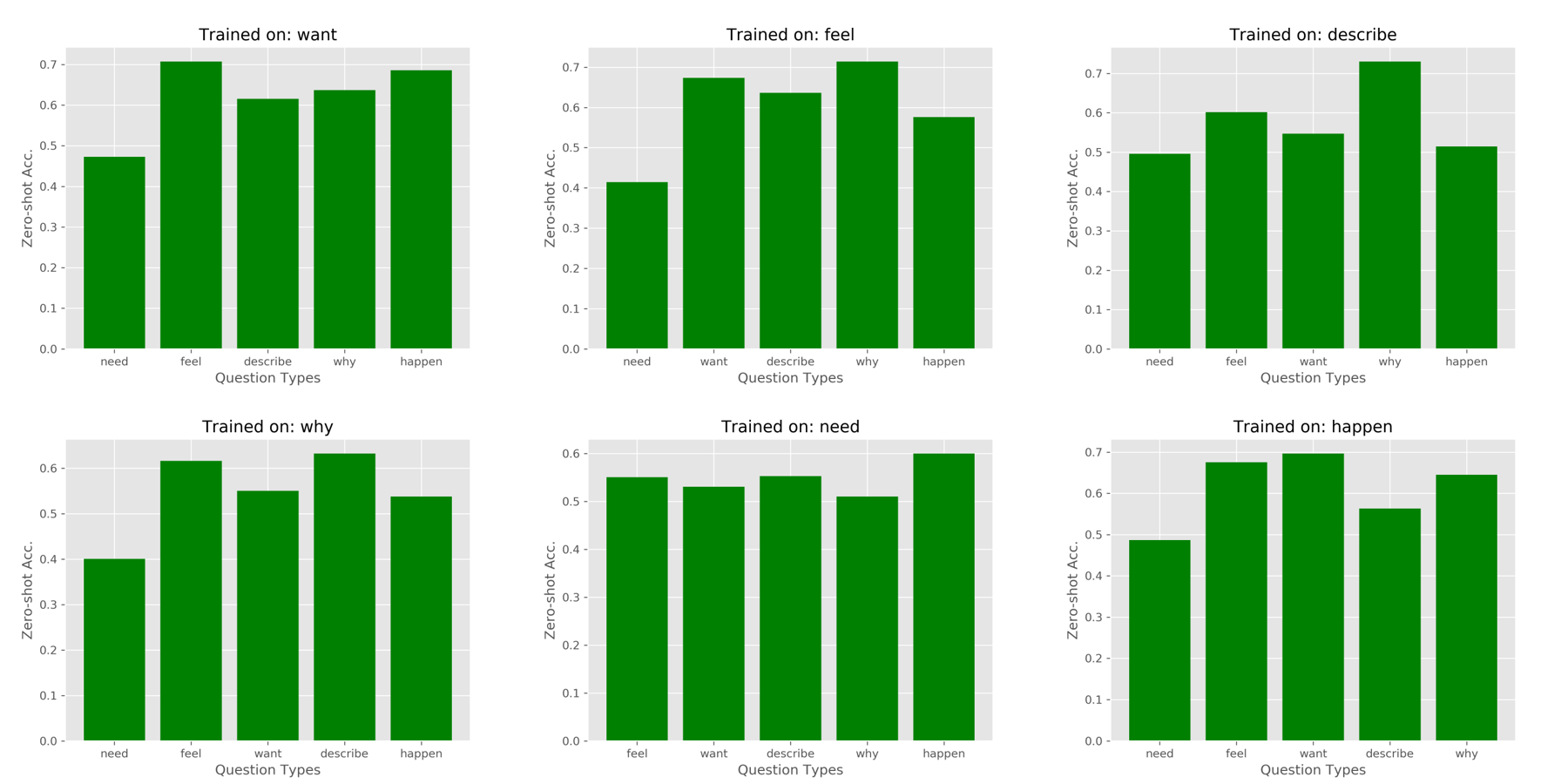}
  \caption{Zero-shot transferring among 6 question types. The top-left type (\emph{want}) accounts for the most training instances, while the bottom-right type (\emph{happen}) has the least. Each bar corresponds a transfer pair.}
  \label{fig:type}
\end{figure*}

\textbf{Question Type Transfer}
Furthermore, we fine-tuned our model based on different question types of training instances\footnote{Instances in Social IQA can be divided into 6 question types, i.e., \emph{wants, reactions, descriptions, motivations, needs, effects.} Since the data does not contain such annotations, here we categorize instances by keyword matching. Our categorization has a similar distribution to that in the initial paper.}, where the model is tuned on only one type and evaluated on the others. This setting challenges the model's ability to do zero-shot question type transferring. The results in Figure~\ref{fig:type} show that most of the pairs are much better than random guessing ($\sim33\%$), even though some types, such as \emph{why, need, happen}, account for less than $15\%$ training instances.

\textbf{Zero-Shot Task Transfer}
The model utilizing additional multiple-choice datasets shows promising results in Table~\ref{table:proposed}. Here, we perform zero-shot transferring between these datasets and Social IQA, showing that the knowledge learned from these tasks indeed is highly transferable to Social IQA (Table~\ref{table:transfer}). Among them, ComsmosQA is the most transferable, which makes sense since it has the same data format of $(C, Q, A)$ as Social IQA, despite having much longer context.

\begin{table}[ht]
  \centering
  \begin{tabular}{|l|c|}
  \hline
	\bf Training Task & \bf Social IQA (\%) \\
  \hline\hline
    HellaSwag & 52.7 \\
    CosmosQA & 63.7 \\
    WinoGrande & 59.2 \\
    \hline
    None & 33.3 \\
  \hline
  \end{tabular}
  \caption{Fine-tuning on three different commonsense tasks and evaluating on Social IQA's dev set.}
  \label{table:transfer}
\end{table}

\textbf{Information Masking}
Previous work~\cite{feng2018pathologies, kavumba2019choosing, trichelair2019reasonable, tamborrino2020pre} on question-answering has shown that some models tend to learn from artificial or superficial patterns of the dataset, and they can still predict the correct answer after import clues (to human) in the premise are masked. Therefore, we challenge our well-trained model, a RoBERTa with MLM fine-tuned on the original training set, by masking context, question, and both, respectively, during inference. We evaluated three different masking methods: 
\begin{itemize}
	\item  Zero-pad the masked segment, so the sequence length remains unchanged.
	\item  Delete the entire masked segment.
	\item  Filter out the keywords in the masked segment, where we define keywords as \emph{noun, verb, adjective}\footnote{We used the POS tagger from \url{https://spacy.io/models/en}} and other task-important words like \emph{``before, next, What, How, Why"}. For example, \emph{``Tracy didn't \underline{go home} that \underline{evening} and \underline{resisted} Riley's \underline{attacks}."} becomes \emph{``Tracy didn't that and Riley's"}.

\end{itemize}

Table~\ref{table:mask} shows that without either context or question, the model's performance does drop significantly, and when both are not provided, it degrades even more. 
The patterns for different masking approaches are the same. Note that \textbf{Filter} may not remove all the useful words, but we can already see the significant performance drops. 
This analysis shows the model is relying on meaningful information to make predictions.

\begin{table}[ht]
  \centering
  \begin{tabular}{lccc}
    \hline
    \multirow{2}{*}{\bf Info.} & \multicolumn{3}{c}{\bf Accuracy (\%)}\\
    \cline{2-4}
     & \bf Pad & \bf Delete & \bf Filter \\
    \hline\hline
    Full & \multicolumn{3}{c}{78.4} \\
    \hline
   - Context & 51.8 & 52.2 & 58.4 \\
   - Question & 58.4 & 57.8 & 64.3 \\
   - Both & 44.3 & 40.0 & 47.7 \\
\hline
  \end{tabular}
  \caption{Effect of different information masking methods on a well-trained RoBERTa model.}
  \label{table:mask}
\end{table}

\vspace{-0.1in}
\section{Conclusion}
In this paper, we attempt to build a system with commonsense social intelligence. 
Building on powerful pretrained language models, we propose several variations focusing on different aspects: alleviating the training instability problems by jointly training the model with the MLM and multiple-choice classification loss, considering Social IQA's properties and designing more sophisticated architectures, utilizing external commonsense resources, and ensembling models with complementary properties. 
Our proposed methods achieve competitive results on Social IQA leaderboard. Besides, we found that the pretraining step highly dominates the performance, and a large pretrained language model such as RoBERTa can generalize commonsense knowledge well. 
Though our experiments are focused on Social IQA, we believe our methods and findings may help future work to develop models for similar tasks and machine social intelligence in general.

\bibliographystyle{IEEEbib}
\bibliography{strings,refs}

\end{document}